# A COMPARISON OF SEMI-DETERMINISTIC AND STOCHASTIC SEARCH TECHNIQUES

A M Connor and K Shea
Engineering Design Centre, University of Cambridge
email: amc50@eng.cam.ac.uk, ks273@eng.cam.ac.uk

**Abstract.** This paper presents an investigation of two search techniques, tabu search (TS) and simulated annealing (SA), to assess their relative merits when applied to engineering design optimisation. Design optimisation problems are generally characterised as having multi-modal search spaces and discontinuities making global optimisation techniques beneficial. Both techniques claim to be capable of locating globally optimum solutions on a range of problems but this capability is derived from different underlying philosophies. While tabu search uses a semi-deterministic approach to escape local optima, simulated annealing uses a complete stochastic approach. The performance of each technique is investigated using a structural optimisation problem. These performances are then compared to each other as and to a steepest descent (SD) method.

## 1. Introduction

The purpose of this paper is to investigate the performance of two optimisation techniques applied to an engineering design problem. Tabu search [1] is an aggressive metaheuristic that guides a local search out of local optima while simulated annealing [2] uses a probabilistic approach to obtain the same end. Comparing heuristic techniques is difficult since performance is highly dependant on the specific formulation of the general method. In this comparison, the implementation of each technique is sufficiently mature such that a direct comparison is unlikely to lead to results that favour one particular method due to bias in development effort. The tabu search algorithm used in this study was originally developed for application to the optimisation of fluid power circuits [3]. The simulated annealing algorithm is the underlying search technique used in an approach to size, shape and topology optimisation of structures [4].

The problem chosen to assess the relative merits of each algorithm is the shape and size optimisation of a ten bar truss. This is a simple extension of the standard problem, which only considers size optimisation, and a step towards applying tabu search to structural topology optimisation.

## 2. Optimisation Techniques

A variety of techniques can be used to tackle the optimisation of engineering design problems. Traditional methods, such as steepest descent and conjugate gradient methods have recently fallen into disfavour due the advent of claimed global optimisation methods such as genetic algorithms, tabu search and simulated annealing. A wealth of studies exist in the literature which compare different methods [5,6,7] although no general conclusions can be made as sample problems come from differing domains and there are different approaches to implementing both the algorithms and the representation used.

## 2.1. Tabu Search

The tabu search concept is a heuristic procedure designed to guide other methods to avoid local optimality. Tabu search has been shown to be effective on a wide variety of classical optimisation problems, such as graph colouring and travelling salesman problems, and has also been applied to practical problems such as scheduling and electronic circuit design. The method uses constraint conditions, such as aspiration levels and tabu restrictions, and a number of flexible attribute based memories with different time cycles. The flexible memories allow search information to be exploited more thoroughly than rigid memory or memory-less systems, and can be used to either intensify or diversify the search to force the method to find optimum solutions.

The underlying search method in the current implementation is a variable step size steepest descent algorithm. This has been chosen as it allows a direct search to be carried out without any gradient information. Two memory lists are used to control the search algorithm. The short term memory contains representations of recently visited solutions that are classed as tabu. When the search algorithm locates an optimum it is forced to make a move and it is not allowed to return to a solution that is contained in the list. As the search progresses the list is updated on a first in-last out basis so that the list remains a fixed size. It is the short term memory and the notion of tabu restriction that provide the capability to escape local optima.

The intermediate term memory is similar to the short term memory, but it contains a list of previously visited best solutions. This list of solutions is used to provide a means of focussing on good regions of the solution space. This intensification provides the means to accelerate the search by examining trends in good solutions and proposing new solutions based on an extension of these trends. In addition to this, intensification can locate new good solutions by investigating the centroid of several disparate good solutions. The final search control mechanism is diversification. This is often implemented by using an additional long term memory cycle but in this implementation is based on a simple refreshment involving a scattering of new random solutions.

## 2.2. Simulated Annealing

Simulated annealing algorithms model the process of annealing in solids to optimise complex functions or systems. In the physical world, annealing is accomplished by heating a solid to an elevated temperature and then allowing it to cool slowly enough so that the thermal equilibrium is maintained. Atoms in the material then assume a globally minimum energy state. Simulated annealing algorithms have been successfully applied to a variety of problems. The algorithm starts with an initial design and generates a new design by changing one or more of the design variables. The objective function is then evaluated for the new design. While a better design is always accepted there is a possibility that a worse design may be accepted based on a probability function [8].

The change in energy is expressed as the change in objective function value, while the temperature is a control parameter that sets the probability of selection. In the general method, the temperature is held constant for a prescribed number of iterations to allow the system to gain "thermal equilibrium" and is then decreased in accordance with a cooling curve. As the temperature decreases, so does the probability that an inferior design will be accepted. This forces the algorithm to converge to an optimal, or near optimal, solution.

Simulated annealing algorithms are reasonably robust if the parameters controlling the cooling curve are assigned values that reflect the complexity of the problem. In this implementation the modified Lam-Delosme schedule is used [9] with rules selected based on quality metrics [10] and dynamic constraint weights [11].

## 3. Optimisation of the Ten Bar Truss

The problem considered in this paper is the optimisation of the ten bar truss. In the standard approach to this problem the spatial layout of the truss is constrained as shown in Figure 3.1 by fixing the position of the nodes.

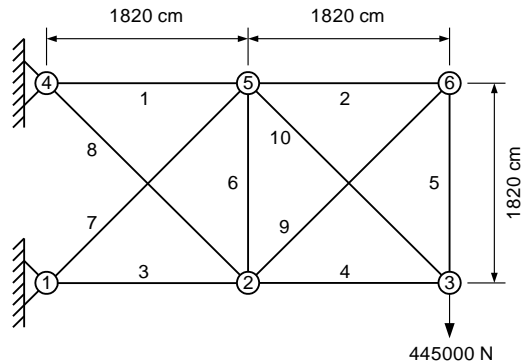

Figure 3.1: Ten bar truss

The truss is assumed to consist of an idealised set of pin jointed bars connected together at the nodes. The design optimisation problem is to find the cross sectional areas of each member such that the mass is minimised. The problem has been expanded in this paper to allow the spatial layout of the truss to be adjusted as well as the cross sectional area of the members. The built in nodes are fixed in position, as is the loaded node. This variation of the problem therefore introduces six new parameters that determine the position of nodes 2, 5 and 6 in Cartesian space. It is important to realise that the introduction of new parameters does not turn the problem into a topology optimisation problem. Even if member areas were allowed to drop to zero, this removal of members is only a reduction in the connectivity of a given topology. A true topology optimisation can only be achieved if nodes can be added and removed.

Essentially, all of the design parameters are continuous. However, the nodal positions have a minimum allowable change of 1cm whilst for the cross sectional areas the minimum allowable change is 0.01cm$^2$. In reality, these cross sectional areas would be limited to discrete values corresponding to available stock material. The material for the truss is aluminium with Young's modulus of 6.88x10$^6$ N/cm$^2$ and material density is 2.7$^{-3}$ kg/cm$^3$. Each member is modelled as a solid circular cross section.

### 3.1. Constraints and Convergence

Despite the simplicity of the ten bar truss example, it is still a reasonably constrained problem due to the difficulty in finding high quality, i.e. low mass, solutions that do not violate either the buckling or stress constraints. The constraints on the problem are that each member should not have a stress that exceeds 17,200 N/cm$^2$, buckle under Euler buckling criteria, and have a length less than 15cm.

The tabu search and simulated annealing implementations used in this paper take different approaches to dealing with constraint violations. In order to reflect the aggressive nature of the search, the tabu search method uses a simple rejection of infeasible solutions. In comparison, the simulated annealing approach used in this study uses a complex, dynamically weighted penalty function that decreases the allowable violation as the search progresses. This approach allows the method to track through regions of infeasibility in order to locate new feasible solutions where as the tabu search implementation relies on the intensification of trends to carry the search through infeasible regions.

Both methods have the capability to test convergence of the search and induce a premature termination of the search. However, the implementations of convergence criteria differ significantly and have not been used in obtaining the results. The implication is that the simulated annealing approach will carry out all of the evaluations specified by the cooling strategy and a potentially large number of the evaluations used by the tabu search will only be producing very small decreases in mass.

## 4. Results

An initial investigation of optimising only member size resulted in solutions in the range 2900 to 3300kg. These results have been used to set a target mass threshold of 2900kg to determine the degree of success of the methods on the expanded problem.

Results are presented for ten optimisation runs of each method. In the first instance, results are presented that show summary of the results for test runs for all three methods. Both the tabu search method and the underlying steepest descent method are initialised by generating a random scatter of solutions with the search being started from the best feasible solution. The simulated annealing method is started from the same point but has an initial random walk which takes it to a different region of the solution space in each run before the probability of accepting a worse solution is reduced below 1. Therefore, the multiple runs of the problem indicate how the methods perform when started from different regions of the solution space. The results summary is shown in Table 4.1.

|  | SD | TS | SA |
|---|---|---|---|
| Best mass (kg) | 2299 | 1598 | 1491 |
| Best mass evals | 4206 | 12004 | 34000 |
| Worst mass (kg) | 7208 | 2948 | 2307 |
| Worst mass evals | 3252 | 8896 | 34000 |
| Average mass (kg) | 4162 | 2401 | 1967 |
| Std. Dev. of mass (kg) | 1796 | 495 | 323 |
| Lowest evals | 2711 | 8441 | 34000 |
| Highest evals | 5676 | 36806 | 34000 |
| Average evals | 4386 | 13455 | 34000 |
| No. runs below 2900kg threshold | 3/10 | 8/10 | 10/10 |

Table 4.1: Statistical comparison of methods

The summary of the results shows the best and worst masses achieved by each method, along with the number of evaluations required in each case. The simulated annealing algorithm always carries out the same number of evaluations where as for the other methods the number of evaluations is dependant on how the search progresses. In addition to this, the summary also shows the average mass of the ten runs and the standard deviation away from that value. This is essentially an indication of the consistency of the performance. Finally, the lowest, highest and average number of evaluations is shown.

Due to the dynamic penalty function used by the simulated annealing approach, a number of solutions found exhibited minor stress violations including the best solution found. Manually adjusting these solutions to remove the violations led to a slight increase in mass and it is the adjusted values used in Table 1. The following results first compare the best initially feasible solution found by the simulated annealing approach to those found by the tabu search and steepest descent methods. The adjusted solution that has the lowest mass is then described.

Table 4.2 shows the numeric values for each of the design parameters where the x,y coordinates of the node positions are expressed relative to the lower fixed node. The solution shown for the simulated annealing approach is the best solution found with no residual constraint violations. The shading in the table indicates minimum area members. Such minimum area members can be removed from the structure to produce a reduced topology solution provided that the change does not produce a violation of constraints or so great a reduction in topology that the structure begins to act as a mechanism. Simple violations may be adjusted out by the designer, however it is important to realise that if a minimum area member is transferring significant force to a node, then removing that node may result in significant changes in the response of other members which become increasingly difficult to gauge.

|  | SD | TS | SA |
|---|---|---|---|
| $x_2, y_2$ (cm) | 488,89 | 445,-61 | 568,-151 |
| $x_5, y_5$ (cm) | 840,581 | 807,408 | -13,920 |
| $x_6, y_6$ (cm) | 1436,-44 | 1197,-112 | 1252,-176 |
| $A_1$ (cm$^2$) | 59.71 | 60.39 | 65.29 |
| $A_2$ (cm$^2$) | 28.17 | 16.6 | 22.92 |
| $A_3$ (cm$^2$) | 204.14 | 183.17 | 242.63 |
| $A_4$ (cm$^2$) | 0.01 | 0.01 | 0.01 |
| $A_5$ (cm$^2$) | 132.31 | 239.9 | 193.82 |
| $A_6$ (cm$^2$) | 125.66 | 3.04 | 13.78 |
| $A_7$ (cm$^2$) | 0.01 | 0.01 | 0.19 |
| $A_8$ (cm$^2$) | 145.94 | 1.42 | 0.01 |
| $A_9$ (cm$^2$) | 374.43 | 310.26 | 259.65 |
| $A_{10}$ (cm$^2$) | 42.52 | 47.9 | 35.66 |
| Mass (kg) | 2299 | 1598 | 1526 |

Table 4.2: Comparison of best solutions

The best truss found by the steepest descent algorithm is shown in Figure 4.1.

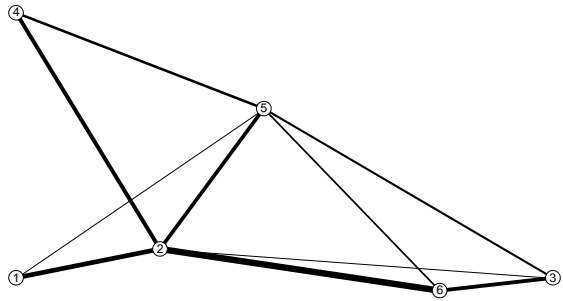

Figure 4.1: SD solution

Removing the minimum area members of the solution shown in Figure 2 produces a minor buckling violation in the member between nodes 2 and 3; see Figure 4.2. Increasing the area of this member from 145.94 cm$^2$ to 146 cm$^2$ removes the violation and makes the reduced topology to be a feasible design. The fact that such a tiny increase in area removes the buckling violation suggests that removing the hard constraints or introducing a tolerance on the constraints may be of benefit as such small violations are insignificant.

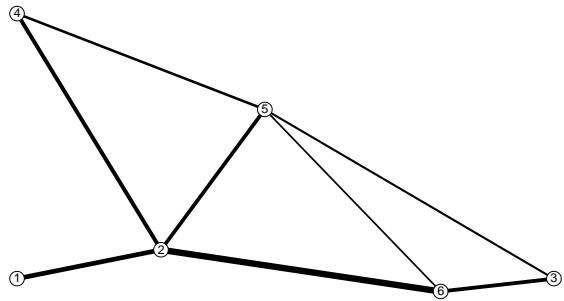

Figure 4.2: Adjusted SD solution

The solution in Figure 4.3 resulted from the tabu search method.

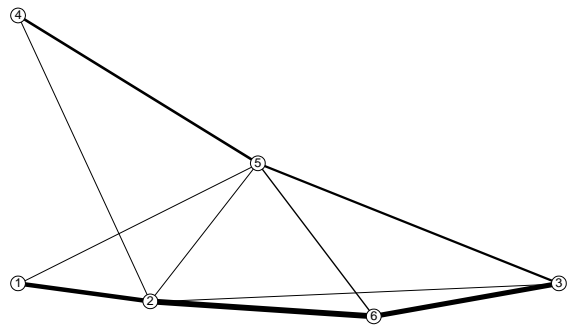

Figure 4.3: TS solution

Removing the minimum area members produces the solution shown in Figure 4.4. The reduced topology has no constraint violations and needs no adjustment.

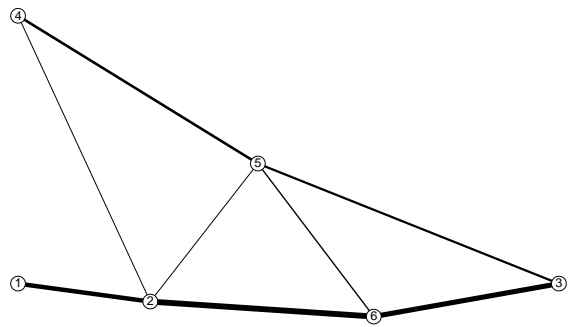

Figure 4.4: Adjusted TS solution

This solution has the same reduced topology as that found using the steepest descent method but is considerably lighter. Due to the different joint positions the solution has smoother load transmission paths as well as a more elegant appearance. Also, the variance of member lengths is noticeably smaller than in the solution found by the steepest descent method.

The solution shown in Figure 4.5 is the best feasible solution found by the simulated annealing algorithm. The position of node 5 has moved just behind the support node since no spatial constraint was formulated. This may cause problems depending on how the truss is supported. The support node, node 4, and the member joining nodes 4 & 5 is obscured in the Figure 4.6 but the length of the member is longer than the minimum length and an expanded view of this region (not to scale) is shown in Figure 4.6.

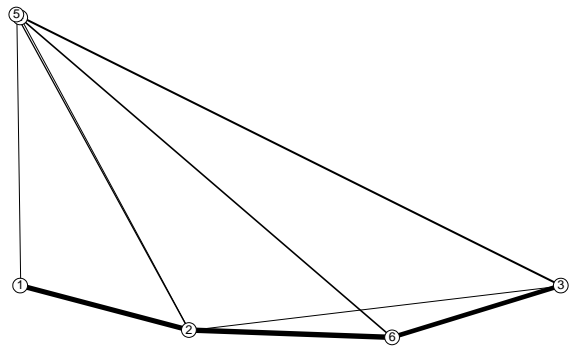

Figure 4.5: Feasible SA solution

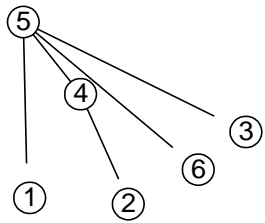

Figure 4.6: Expanded region of solution

From this expanded view it can be seen that the positioning of node 5 above and behind the fixed support node has resulted in a compression member between nodes 4 & 5 which transmits the forces applied to node 5 from the other members to the support.

Removing the minimum area members does not produce violations. Although it is not shown, it is also possible to remove member 7 which has a small area, .19 cm$^2$ with no adverse impact on the behaviour of the structure. It may also be possible to remove node 5 and attaching the tension members that link node 5 to nodes 6 & 3 directly to the support node. This has the benefit of removing the compression member between nodes 4 & 5 which is a potential source of buckling failure. These potential changes have not been incorporated in the adjusted solution that is shown in Figure 4.7 even though they would result in a simplified reduced topology solution.

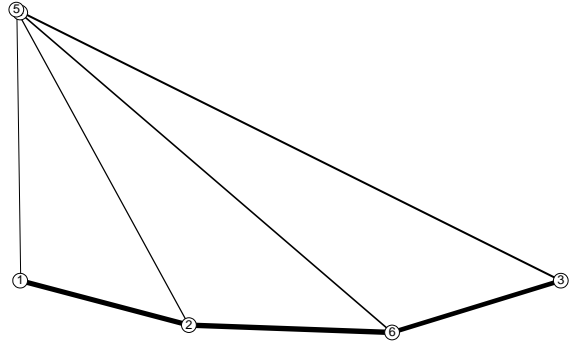

Figure 4.7: Adjusted feasible SA solution

The simulated annealing approach did locate a number of solutions that were slightly infeasible due to the soft constraint formulation. The best solution found overall after small violations were removed has the parameter values given in Table 4.3.

|  | SA |
|---|---|
| $x_2, y_2$ (cm) | 612,-92 |
| $x_5, y_5$ (cm) | 93,850 |
| $x_6, y_6$ (cm) | 1259,-96 |
| $A_1$ (cm$^2$) | 61.36 |
| $A_2$ (cm$^2$) | 11.64 |
| $A_3$ (cm$^2$) | 252.61 |
| $A_4$ (cm$^2$) | 0.01 |
| $A_5$ (cm$^2$) | 202.02 |
| $A_6$ (cm$^2$) | 8.05 |
| $A_7$ (cm$^2$) | 0.01 |
| $A_8$ (cm$^2$) | .74 |
| $A_9$ (cm$^2$) | 251.89 |
| $A_{10}$ (cm$^2$) | 43.47 |
| Mass (kg) | 1491 |

Table 4.3: Initially infeasible SA solution (adjusted to become feasible)

The minimum area members, four and seven, were removed whilst the area of member ten was altered from 28.41 cm$^2$ to 43.47 cm$^2$. The area of member six was slightly altered from 8.03 cm$^2$ to 8.05 cm$^2$ to obtain this solution. This solution is shown in Figure 4.8.

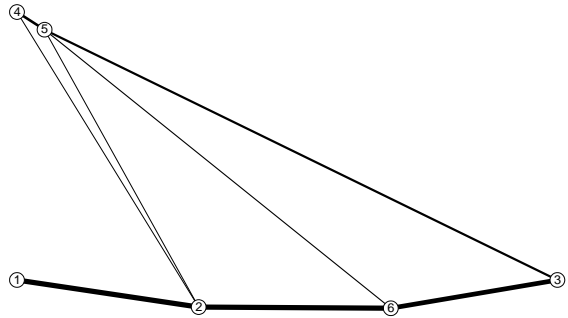
Figure 4.8: Adjusted SA solution that initially had violations

## 5. Discussion

A goal of this study was to assess the merit of each method in light of the computational expense required and the quality of the solution obtained. Using steepest descent as the basis, the best solution found by tabu search is 31% lighter with 2.85 times the number of evaluations. Simulated annealing resulted in a solution that is 35% lighter but required eight times the number of evaluations. While for the ten bar truss problem the evaluation is not highly time intensive, for larger problems and situations where the evaluation is more computationally expensive, reducing the number of evaluations is highly desirable. The expense of the optimisation must also be compared to the number of designs that will be manufactured and additional cost benefits outside of material purchase for reducing the mass of the structure.

An advantage of tabu search is a semi-deterministic nature as it acts both as a local and global search method. With the current implementation, from an initial starting point, tabu search always results in the same solution. It was seen though that the performance of tabu search improved by using an initial scatter search but, this eliminated the deterministic nature. A problem with non-deterministic methods is, while solution quality is improved overall, this can only be assessed based on a number of optimisation runs. A common rule of thumb used with simulated annealing is to take the best solution found from three runs.

The steepest descent and tabu search methods employed a hard constraint for stress and buckling, that is the constraints were never violated at any point in the search. In contrast, simulated annealing used dynamic penalty functions such that constraint violations were allowed throughout the search but were penalised in the cost function. One clear advantage to using soft constraints is that the other methods suffered from simple rejection of solutions with small violations, eg less than 0.01. Considering that the limits modelled in the problem are not accurate to this precision as well as the fact that the structure will not behave exactly as modelled, rejecting solutions with small violations to this accuracy is impractical. This could be resolved using a tolerance on the violations.

As both methods are heuristic a number of search parameters are required. To have a search technique that is robust over a wide range of design problems it is advantageous to reduce the number of parameter adjustments required from problem to problem. The simulated annealing implementation used has 6 temperature schedule parameters, 15 parameters associated with the selection of moves, and 8 parameters associated with the dynamic constraint weights. While the majority of these parameters are robust over a wide range of problems, some fine tuning is generally necessary. In comparison, the tabu search method has a total of 5 control parameters for which an empirically derived heuristic has been developed that relates the values required to the number of design parameters.

Many structural optimisation methods allow members to reduce to a minimum area, which is generally very small, with the implication that these members can be removed from the structure. As was shown in Section 4, this is not always a simple task. Skill and understanding are necessary to transform the solutions found by the optimisation algorithms into sensible designs. Deciding which members can be removed without the structure collapsing or inducing stress and buckling violations is non-trivial. This emphasises that optimisation can only be an aid for designers and engineering judgement is still imperative.

To reduce the mass of the structure farther, topology changes are required. Simulated annealing has been applied to this problem in the past and a best mass of 853kg was reported for the same problem [12]. A goal of this study was to investigate the advantages of tabu search over simulated annealing for topology optimisation of discrete structures. General advantages are a decreased number of evaluations, potential for parallel evaluation, fewer search parameter adjustments, possibilities for incorporating design knowledge, and potential for learning. These advantages could lead to a more appropriate search technique for topology design problems.

## 6. Conclusions

This paper compares the performance of two mature implementations of different search algorithms to structural optimisation. In general, both methods perform considerably better than a local search method, with simulated annealing finding higher quality solutions than tabu search at the cost of increased computational expense.

Each of the methods has some advantages and disadvantages other than improved solution quality and reduced evaluations. The aim of future work is to combine strengths of each method in such a to produce a hybrid. This may involve the use of memory cycles to escape optima and incorporate domain knowledge and utilise search history. The use of probabilistic moves will be investigated so as to remove the inherent disadvantage of tabu search in searching all moves that penalises the current implementation when the number of variables is very large. The intent of the resulting method is an efficient and effective technique for general engineering topology problems.